# A Dynamic 3D Spontaneous Micro-expression Database: Establishment and Evaluation

Fengping Wang, Jie Li, Siqi Zhang, Chun Qi, Yun Zhang, Danmin Miao

**Abstract**—Micro-expressions are spontaneous, unconscious facial movements that show people's true inner emotions and have great potential in related fields of psychological testing. Since the face is a 3D deformation object, the occurrence of an expression can arouse spatial deformation of the face, but limited by the available databases are 2D videos, lacking the description of 3D spatial information of micro-expressions. In this paper, we proposed a new database that contains 373 micro-expression samples, each consisting of 2D video sequence and corresponding 3D point cloud sequence. These samples were classified using objective method based on facial action coding system, as well as non-objective emotion classification method combining video contents and participants' self-reports. We extracted 2D and 3D features using the local binary patterns on three orthogonal planes (LBP-TOP) and curvature algorithms, respectively, and evaluated the classification accuracies of these two features and their fusion results with leave-one-subject-out (LOSO) and 10-fold cross-validation. Further, we used various advanced neural network algorithms for database evaluation, the results show that classification accuracies are improved by fusing 3D features than using only 2D features. The database offers original and cropped micro-expression samples, which will facilitate the exploration and research on 3D spatio-temporal features of micro-expression.

———————————— ◆ ————————————

## 1 INTRODUCTION

Micro-expression is a brief and fast facial muscle movement that reveals genuine emotions that a person tries to conceal [1], [2], [3]. Compared with general facial expressions (also referred to as macro-expressions), micro-expressions tend to occur when people are unconscious, reflect their true thoughts and motivations, and are more likely to occur in high-risk environments. Micro-expressions are not only of high reference significance in lie detection, but are also closely related to inner human emotions and can accurately assess psychological states. Therefore, it has potential applications in national security, justice system, clinical medical, and political election [4], [5].

The short duration of micro-expressions is the main feature that distinguishes micro-expressions from general facial expressions, and researches suggest that the generally accepted upper limit of during is 1/2s [3], [6]. Also, the occurrence of micro-expressions is characterized by low intensity and localization, so it is generally difficult for people to detect or notice them with the naked eye. Even for the well-trained people, the recognition rate is less than 50% [7], [8]. Its characteristics also make the detection and recognition of micro-expressions more difficult and challenging than macro-expressions.

Since the study of micro-expressions in computer vision has only gained attention in the past few years, research on micro-expressions and the publicly available databases of spontaneous micro-expressions used for research are still under development. Based on the existing database, although deep learning methods can achieve more than 80% recognition rate [9],[10],[11], the recognition rate is mostly the result of merging and reorganizing several databases or macro-expression transfer, limited by database size and categories imbalance [12]. Moreover, these databases are 2D static images or dynamic videos, ignoring the 3D depth information of facial micro-expressions. This is insufficient for representing some subtle confusable motions or depth motions that exist on 3D space. Since the human face is a deformable 3D object in space, spatial 3D information is crucial for the description of facial motion. In macro-expression research, several available 3D macro-expressions databases have been established and it has been verified that 3D data play an import role in expression recognition, capable of portraying changes in minute facial features and compensating for some 3D spatial features that cannot be represented by 2D data [13], [14], [15]. The existing micro-

————————————————
- *The work was supported by the National Natural Science Foundation of China under Grant 61675161, Grant 41801250 and Grant 61572395, by the Fundamental Research Funds for the Central Universities under Grant zdyf2017003. (Corresponding author: Jie Li)*
- *F.P.Wang is with the School of Information and Communications Engineering, Xi'an Jiaotong University, Xi'an, China. E-mail: fengpingwang@ stu.xjtu.edu.cn.*
- *J. Li is with the School of Information and Communications Engineering, Xi'an Jiaotong University, Xi'an, China. E-mail: jielixjtu@xjtu.edu.cn.*
- *S.Q.Zhang is with Xi 'an Modern Control Technology Research Institute, Xi 'an , China. E-mail: zsq77bm@163.com*
- *C. Qi is with the School of Information and Communications Engineering, Xi'an Jiaotong University, Xi'an, China. E-mail: qichun@mail.xjtu.edu.cn.*
- *Y.Zhang is with Xi'an Singularity One Information Technology Co., Xi'an, China, E-mail: tvsunny@foxmail.com.*
- *D.M.Miao is with Department of Psychology, Air Force Military Medical University, Xi'an China. Email: psych@fmmu.edu.cn.*

expression data are missing the depth motion and change characteristics of the face in spatial 3D. Therefore, it is essential to create a micro-expression database containing 3D spatial data to investigate more effective feature extraction algorithms by mining facial spatial information.

Based on the above consideration, this paper creates a new database, DSME-3D, to study the variation characteristics of micro-expressions in 3D space. Because the database contains 2D and 3D spatial information of facial expressions, as well as changes of the information in time dimension, the database is a dynamic 3D database (or 4D database). In this paper, the database was labeled and classified using two different approaches, such as expression labeling based on objective AU and emotion labeling incorporating subjective emotions. In baseline evaluation experiments, a comprehensive evaluation of the database using multiple algorithms were conducted to verify that the 3D features can improve the recognition performance of micro-expressions.

The paper is organized as follows. In section 2, we review previous macro-expression and micro-expression databases, and analyze state-of-the-art feature extraction algorithms. In section 3, we introduce the basic information of the DSME-3D database, including the expression elicitation process, sample selection, coding, and contents instruction of the database. Section 4 describes the feature extraction algorithms used for classification estimations and the analysis of experiment results. Section 5 discusses the limitation of the DSME-3D database. Section 6 is the conclusion.

## 2 RELATED WORK

### 2.1 Current Expression Databases

A complete facial expressions database is crucial to research facial expressions. To date, many algorithms have been developed to automatically detect and recognize human facial macro-expressions. Unlike micro-expressions, these expressions are easily noticed and usually last more than half a second and up to four seconds. Numerous databases, such as JAFFE, Multi-PIE, Genki-4K, MMI, CK, GEMEP, Bosphorus, etc. [16], have been published. Among these databases, the samples in the first three databases contained only one facial expression image representing different emotional states. In contrast, the last four databases are dynamic video sequences that can distinguish different expressions better than images. However, these databases are non-spontaneous expression databases that do not reflect people's real emotions. To address this problem, more spontaneous expressions databases were created: CK+ [17], DISFA [18], AM-FED [19], NVIE [20], BU-3DFE [13], BU-4DFE [21], BP4D [22], BP4D+ [23]. These databases have promoted expression research development and made the technology of expression detection and recognition more and more mature. Particularly, with the establishment of the Bosphorus, BU-3DFE, BU-4DFE, BP4D and BP4D+ databases, the study of the 3D spatio-temporal variation features of facial expressions has attracted many scholars. In recent years, the 3D space-based expression feature extraction algorithms have greatly improved the detection and recognition performance of expressions.

### 2.2 Current Micro-expression Databases

As the research of micro-expressions in computer vision has gradually attracted people's attention in recent years, the existing micro-expression databases are relatively limited. So far, there are several published micro-expression databases: USF-HD [24], Polikovsky's Database [25], SMIC database [26], CASME database [27], CASME II database [28], CAS(ME)² database [29], SAMM database [30], MMEW database [31]. The main features of these databases are detailed in Table 1.

USF-HD database [24] includes 100 posed micro-expression samples with a frame rate of 30fps and a resolution of $720 \times 1280$. The participants were asked to watch micro-expression videos, then imitated to make micro-expressions, but the micro-expressions were not clearly labeled, and some expressions lasted longer than 1/2s. Polikovsky's database [25] also includes posed expressions, in which 10 participants posed six types of expressions with a high-speed camera of 200fps and then returned to neutral expressions quickly. These expressions were labeled according to the facial action coding system (FACS). However, micro-expressions occur spontaneously and unconsciously, and the two databases do not effectively identify micro-expressions in the natural state, and the databases have relatively limited sample size.

The SMIC database [26] consists of three datasets, of which the HS dataset was recorded by a high-speed camera of 100 fps and $640 \times 480$ pixels, containing 164 micro-expression clips from 16 participants. The SMIC database did not label action units (AUs), and the labels for expressions were obtained by two coders based on participants' statements about their subjective emotional changes while watching the video, and were divided into three categories: positive, negative, and surprise.

Compared with SMIC, the Institute of Psychology of the Chinese Academy of Sciences has collected a more comprehensive and data-rich micro-expression database, CASME. The CASME database [27] consists of two datasets with a total of 195 micro-expression clips from 19 participants, both of which were recorded using a camera with 60 fps and resolution of $1280 \times 720$, $640 \times 480$. The CASME database labeled facial AUs, as well as the position of the onset, apex, and offset frames when the expression changes. Expressions were classified into seven basic categories based on video content, participants' self-reports, and basic emotion theory.

CASME II database [28] is an improved version of CASME with a higher frame rate (200fps) and face resolution that can capture more detailed facial motion information, and is one of the most widely used databases. The CASME II includes 247 micro-expressions clips from 26 participants. Similar to SMIC, CASME II also elicits micro-expressions from participants through induction. The database was labeled by AUs, participants' self-reports, and movie content. Expressions are classified into five main

TABLE 1
The current micro-expression databases

| Databases | Participants | Numbers | Classes | FPS | Mean Age(SD) | Posed or Spontaneous | FACS Coded |
|---|---|---|---|---|---|---|---|
| USF-HD | N/A | 100 | 6 | 30 | N/A | Posed | N/A |
| Polikovsky | 10 | 42 | 6 | 200 | N/A | Posed | Yes |
| SMIC-HS | 16 | 164 | 3 | 100 | 26.7(N/A) | Spontaneous | No |
| CASME | 19 | 195 | 7 | 60 | 22.03(1.60) | Spontaneous | Yes |
| CASME II | 26 | 247 | 5 | 200 | 22.03(1.60) | Spontaneous | Yes |
| CAS(ME)$^2$ | 22 | 57 | 4 | 30 | 22.59(2.20) | Spontaneous | Yes |
| SAMM | 32 | 159 | 7 | 200 | 33.24(11.32) | Spontaneous | Yes |
| MMEW | 36 | 300 | 7 | 90 | 22.35 | Spontaneous | Yes |
| **DSME-3D** | **33** | **373** | **6/4** | **60** | **25.58(4.91)** | **Spontaneous** | **Yes** |

categories that differ from the common emotional categories.

However, the above databases are mainly used for micro-expression recognition research. CAS(ME)$^2$ and SAMM databases are developed to study the automatic spotting algorithm of micro-expressions. CAS(ME)$^2$ database [29] includes two parts: one part is 87 long videos containing spontaneous macro-expressions and micro-expressions; the other includes 300 cropped macro-expressions samples and 57 micro-expressions samples. CAS(ME)$^2$ database samples were recorded by a camera with a frame rate of 30 fps and a resolution of $640 \times 480$ pixels, and the expressions were labeled in the same way as CASME II.

SAMM database [30] is the first high-resolution dataset that includes 159 micro-expression sequences from multiple ethnicities. Unlike the way all previous databases were elicited and coded, each emotional stimulus video was tailored to each participant, and expressions were objectively classified into seven basic emotional categories based on FACS. Since CASME II and SAMM databases have all the criteria for micro-expressions recognition: emotional categories, high frame rates, rich sample number, and FACS-based movements coding [32], they have become the focus of researchers' attention.

MMEW database [31] is a video-based facial micro-expression database that includes 300 samples from 36 participants. The database also contains 900 macro-expression sequences, which can be used for micro-expression spotting. These samples were recorded using a camera with a frame rate of 90 and a resolution of $1920 \times 1080$, and the expressions were labeled in seven categories.

Although so many databases have been established to study micro-expressions, these databases contain only 2D information about expressions, lacking spatial information about expression changes, which hinders the study and utilization of 3D spatial features of micro-expressions. Existing researches on macro-expressions show that 3D spatial information can capture spatial features missing from 2D data, further improving the detection and recognition accuracies of facial expressions, which motivates us to build the database containing dynamic, spontaneous 3D spatial data of micro-expressions.

## 2.3 State-of-the-Art Feature Extraction Methods

The research based on micro-expression databases mainly focuses on expression recognition, and the quality of features is directly related to the performance of expression recognition. Currently, there are three main types of micro-expression feature extraction algorithms: LBP-TOP, Optical Flow (OF), and other algorithms.

LBP-TOP is an appearance-based feature extraction algorithm that is widely used in micro-expressions recognition. In SMIC, CASME II, and SAMM databases, the basic evaluations are also obtained through the LBP-TOP descriptors [26], [28], [29]. To improve the deficiencies of the original algorithm, many researchers have proposed improvement algorithms based on LBP-TOP. Wang et al. [33] presented LBP with six intersection points (LBP-SIP) volumetric descriptor based on the three intersecting lines crossing over the center points, which eliminated the redundancy of central point reuse of the original LBP-TOP and improved the recognition rate. Zong et al. [34] devised a more general scheme for hierarchical spatial partitioning. The image is divided into multi-scale grids and then weighted the grid features at different scales, so that different expressions have the most suitable grid division to extract feature representations. Huang et al. [35] proposed a new framework based on spatiotemporal face representation to analyze subtle facial movements. LBP operator is used to extract the appearance and motion characteristics of horizontal and vertical projections. However, it ignores the shape attribute of the face image and the discriminative information between the two classes of micro-expressions. Therefore, Huang et al. [36] also proposed an improved distinguishable spatiotemporal LBP algorithm. Yu et al. [37] proposed the Local Cube Binary Pattern (LCBP) algorithm. In addition to acquiring LBP features in a three-orthogonal plane, they used eight templates to convolve $3 \times 3 \times 3$ region of the expression sequence. The local direction microstructure is represented by taking the local max-min directional response and the amplitude information in two directions. Guo et al. [38] proposed an efficient and robust extended LBP-TOP descriptor (ELBPLOP) for micro-expression recognition. It adds two complementary binary descriptors: radial difference LBP-TOP(RDLBPTOP) and angular difference LBP-TOP (ADLBPTOP), to explore the local second-order information along the radial and angular directions contained in the micro-expressions video sequences.

Optical flow (OF) is used to infer object motion by detecting pixel changes between two frames images and is also widely used in micro-expression recognition research. Liu et al. [39] proposed the main directional mean OF (MDMO) feature for micro-expressions recognition. The facial region was divided into regions of interest (ROIs) based on the AUs, and the motion features were extracted by the robust OF method. Later, Liu et al. [40] proposed sparse MDMO.to learn an effective dictionary from a micro-expression video dataset. To emphasize the importance of each motion region, Liong et al. [41] proposed an optical strain-weighted feature extraction algorithm. The algorithm mainly obtains the motion information from optical strain, and then it is pooled spatiotemporally to obtain the weights of different blocks to weight the LBP-TOP features. It improves the distinguishability among different micro-expression classes. Similarly, Liong et al. [42] also proposed a new feature extraction algorithm, the Bi-weighted Oriented Optical Flow (Bi-WOOF), which extracts discriminative weighted motion features only through the onset and apex frames of sample sequences.

Deep learning methods have been used to study micro-expressions in recent years and have achieved better results. Liong et al. [43] proposed OFF-ApexNet to enhance the optical flow features of apex frame and complete the classification of expressions. Monu et al. [44] used dynamic representation of micro-expressions to preserve the facial motion information of a video in a single frame. They also proposed Lateral Accretive Hybrid Network (LEARNet) to capture the facial micro-level features of dynamic representation. Xia et al. [45] proposed a micro-expression recognition method based on deep recurrent convolutional network to get the spatial-temporal deformation of micro-expressions. Zhao et al. [46] proposed a two-stage 3D CNN method from few-shot learning. In the prior learning stage, the generic features of micro-expressions were learned, and advanced features were learned using Focal Loss in the target learning stage. There are also many algorithms based on 3D CNN. Zhao et al. [47] proposed a spatio-temporal graph convolutional network - STA-GCN. The algorithm uses the spatial graph and activation probabilities graph between AUs for feature learning. Since all the existing micro-expression databases are small, they are not suitable for the deep learning required abundant data. In addition to network algorithms for small samples, there are also some multi-database combination or transfer learning algorithm.

Besides, many other feature extraction methods have been developed, such as tensor analysis, sparse representation, deep learning, and other methods. Wang et al. [48] proposed a recognition algorithm based on discriminant tensor subspace analysis (DTSA) and extreme learning machine (ELM). DTSA generates discriminative features from grayscale face images expressed as second-order tensors, further improving ELM classification performance and significantly increasing the recognition rate of micro-expressions. Wang et al. [49] proposed a sparse tensor canonical correlation analysis (STCCA) representation method, which provides a solution to alleviate the sparseness of spatiotemporal information in micro-expressions sequences.

Although micro-expressions recognition researches have achieved relatively better results from traditional algorithm methods to deep learning techniques, it is still insufficient for practical applications. Existing feature extraction algorithms still have many problems. For example, due to the subtleness of micro-expressions, the extracted features are often dramatically affected by changes in personal appearance and facial movements that are irrelevant to expressions [50]. There is a large amount of redundant information in extracted features, and there are still no effective feature extraction algorithms for some categories that are easily confused and difficult to distinguish. To some extent, these problems are limited by the database itself, which also further motivated us to build a micro-expression database containing 3D spatial data to extract more effective features of micro-expressions.

## 3 Dynamic 3D Micro-expression Database

### 3.1 Participant and Equipment

Forty-nine participants were recruited, of which thirty-three for the experimental analysis contained twelve male and twenty-one females, whose mean age was 25.58 (standard deviation = 4.91). Before collection, we consulted the participants about the public database and signed an agreement.

The camera used to capture 3D facial expressions is Intel RealSense D415 camera, which contains an RGB camera and depth camera that can capture accurate color images and corresponding deep or point clouds data of the target. The video samples were collected in a compressed file, containing color sequences, deep sequences, and point clouds sequences.

### 3.2 Elicitation Material and Procedure

Using video episodes with high emotional valence is the key to eliciting micro-expressions [28]. Referring to the existing micro-expression database, we used some standard film from Gross [51] and from the Asian Emotion Video library [52], and added some videos from the web. Finally, fourteen video clips involving five basic expression categories, such as happiness, anger, sadness, surprise, and disgust, were used as elicitation materials. The length of these episodes ranged from 41 seconds to 3 minutes. The emotional classes of materials were based on the subjective ratings of thirty-nine volunteers, that is, more than half of the volunteers have same emotion for the same videos. Of course, there are individual differences in subjective feelings about the same video clip, and a clip may elicit several different emotions, so participants only selected one as the main emotion evoked by the video. These participants scored intensity of the main emotions elicited by these videos on a scale of 1 to 5, where 1 represents the weakest and 5 represents the strongest. Details of the video clips are shown in Table 2.

The sample collection environment is the laboratory, using two professional LED lamps for lighting to avoid flickering as much as possible. The camera is fixed above the computer, directly in front of the people's faces. The resolutions of both the color and depth cameras were set to $848 \times 480$ pixels, and the frame rate is 60fps (see Fig. 1).

TABLE 2
Video materials for eliciting micro-expressions

| NO. | During | Main Elicited Emotion | Rate of Selection | Mean Score |
|---|---|---|---|---|
| 1 | 1'33" | Sadness | 0.92 | 3.19 |
| 2 | 2'35" | Sadness | 0.95 | 4.50 |
| 3 | 3' | Sadness | 0.81 | 2.86 |
| 4 | 41" | Anger | 0.82 | 4.62 |
| 5 | 1'54" | Anger | 0.79 | 3.52 |
| 6 | 1'31" | Happiness | 0.77 | 2.67 |
| 7 | 1'38" | Happiness | 0.72 | 3.00 |
| 8 | 2'45" | Happiness | 0.92 | 3.42 |
| 9 | 2'8" | Surprise(disgust) | 0.54(0.36) | 2.73(3.43) |
| 10 | 1'37" | Surprise | 0.74 | 2.65 |
| 11 | 1'27" | Surprise | 0.92 | 2.47 |
| 12 | 1'34" | Surprise | 0.79 | 3.42 |
| 13 | 42" | Disgust | 0.95 | 4.24 |
| 14 | 1' | Disgust | 0.97 | 4.68 |

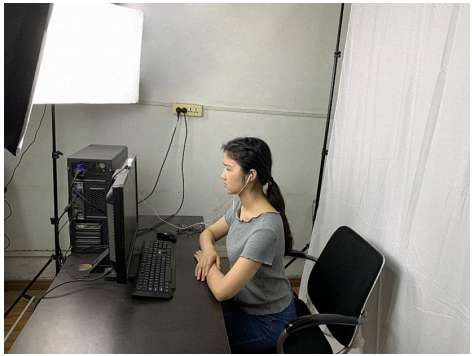

Fig.1. Emotion-collecting environment.

To maximize the ability to suppress expression naturally, we cleared out the other researchers around the areas to make participants as comfortable as possible. We asked participants to maintain a neutral expression while watching the video, and to try to suppress their facial movements when they felt a relevant expression [2], [28]. During viewing the video, it is required to avoid head movements and stare at the screen as much as possible and watch the movie carefully [29]. Participants could also ask to stop watching the video at any time, as the video might overstimulate their emotions and cause the process to be aborted [30]. To avoid the interaction of different video materials on participants' emotions, each video was watched and the next video was started after participants' emotions had calmed down. To better elicit hidden expressions, participants were also informed before starting that prize rewards would be offered for good performance, and that all videos' contents would be kept confidential after watching videos.

The collection begins with a staff member activating the camera. Participants watched the video material, and the sample was saved after each video viewing. After viewing all the videos, participants were asked to provide inner feelings on the facial movements produced during the videos, which was used to aid in the coding of facial expressions.

### 3.3 Sample Selection

Two well-trained coders analyzed the samples frame-by-frame to determine where the micro-expressions appear, how long they lasted, and label the location of the onset frame, apex frame, and offset frame. The captured video was processed as follows.

First of all, since the original files are compressed files, it is necessary to decompress them to get color sequences and point clouds sequences of video.

Secondly, the coders browsed the videos, roughly located all the facial micro-movement's locations present in the video, and saved clips that last less than one second as candidates.

Then, they viewed the candidate clips repeatedly, determined the exact onset and offset frames of each movement clip using the frame-by-frame method. According to the timestamps of the sequence frames, the samples that satisfy the micro-expression duration condition, the total duration less than 500ms or the onset duration less than 260ms, are used as final micro-expression samples. Samples with micro-expressions too subtle to be coded were removed [28], some samples with accompanying blinks but with expressions were retained.

Finally, after the color sequences of micro-expressions were determined, the point clouds sequences were extracted correspondingly according to the frame timestamps. Fig.2 shows the 2D and the corresponding 3D point clouds reconstruction sequence frames.

### 3.4 Coding and Categories Labeling

The current database classification methods are different. FACS is an objective method to label facial movements with action units [53]. Since strong emotional stimuli can elicit spontaneous micro-expression, some studies believed that it is inappropriate to forcefully classify micro-expression into six categories that are identical to ordinary facial expressions. Therefore, in CASME II and CAS(ME)² database, the expressions were labeled by combining FACS, participants' self-report, and video contents [28], [29].

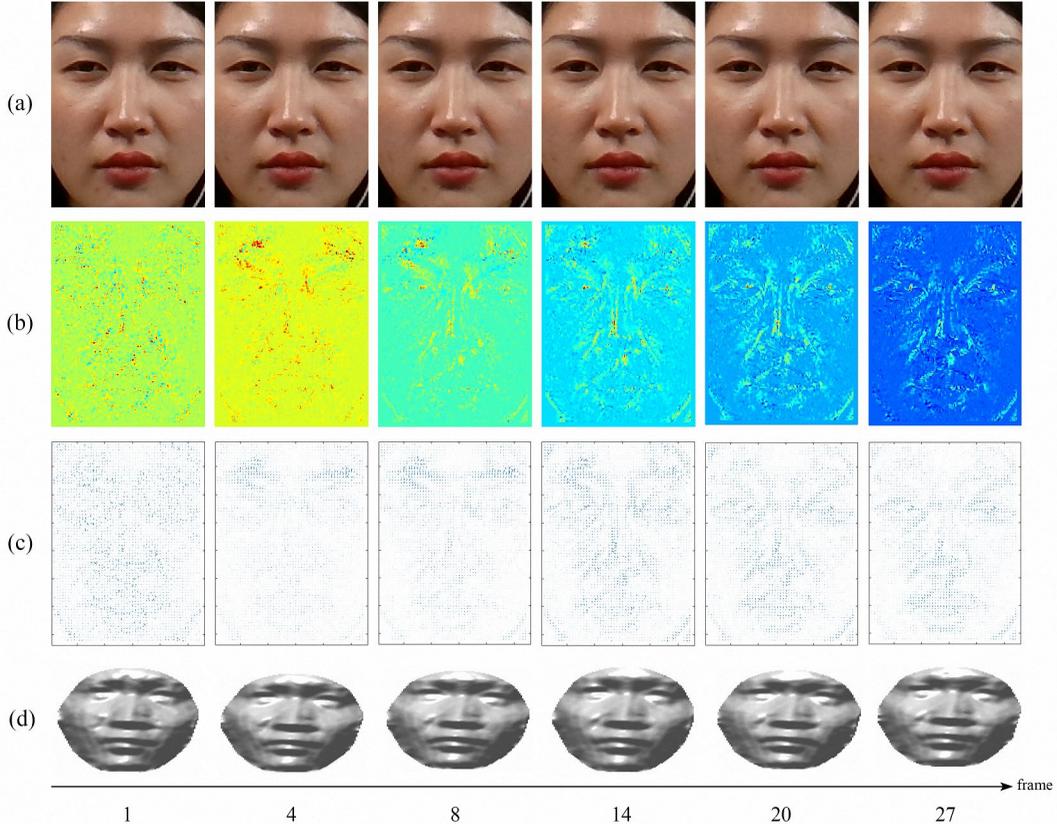

Fig.2. Micro-expression samples. The movement is shown in AU1+2, apex at frame 8. (a) 2D sequences, (b) optical flow fields, (c) optical flow directions, (d) corresponding to 3D point cloud reconstruction sequences frames.

However, some studies currently held that the movements are objective, and it would be a more reasonable method to use FACS to guide emotion classification [3]. In the SAMM database coding, only FACS was used to provide information about emotions [30]. In 2017, Davison et al. [54] addressed the FACS coding method. They believed that the same movements in different categories have some effect during machine learning, and proposed new classes based on FACS to carry out reclassification tests on CASME II and SAMM database. Although recognition rates have improved, there is no suggestion that the kind of labeling is reasonable.

Due to the localized nature of micro-expressions, the association between facial movements and felt emotions has not been clearly defined [2]. For the same emotion, different people have different facial movements. And the representation of emotions should not only include facial movements, but should also take into account the psychological and physiological reactions of the participants. If only consider facial micro-expression and not emotional types, just through FACS. In this paper, two well-trained coders who obtained coding certificates coded AUs and la-beled objectively samples for micro-expression classification. Further, AUs, video content and self-reports were combined to classify the samples in order to analyze the corresponding emotional categories of micro-expressions. Among them, the AUs caused by blinking and eye rolls were not coded, and samples for which the emotional category could not be determined were recorded as "Others". Table 3 and Table 4 list the sample number and standard AUs of the two classified emotions in the database.

The reliability between two coders in the database is 0.84, which was calculated as

$$R = \frac{2 \times AU\left(C_1, C_2\right)}{All_{AU}} \tag{1}$$

where $AU\left(C_1, C_2\right)$ is the number of AUs on which Coder 1 and Coder 2 agreed, $All_{AU}$ is the total number of AUs in a micro-expression scored by the two coders. Afterwards, the two coders discussed and arbitrated the disagreements [53].

## 3.5 DSME-3D Database User Guide

The DSME-3D database samples are spontaneous, dynamic sequences of micro-expression. This database consisted of 373 micro-expressions from 33 subjects, each containing color sequence frames and corresponding point clouds sequence frames. These 2D and 3D sequence frames are corresponded one-to-one by timestamps. Each sample was labeled with onset frame, apex frame, offset frame, and the corresponding AUs. The AUs labeling of the micro-expressions is based on FACS Investigator's Guide and The Manual [53], [55]. The classification categories contain objective classification and non-objective classification of micro-expressions. Because some types of micro-expressions are difficult to elicit under laboratory conditions, the distribution of samples in different categories is uneven.

TABLE 3
AU labeling and frequency for database objective classification

| Classes | Number | AUs |
|---|---|---|
| Happiness | 39 | AU6, AU12, AU6+AU12, AU6+AU7+AU12, AU7+AU12 |
| Surprise | 36 | AU1+AU2, AU5, AU25, AU1+AU2+AU25, AU25+AU26, AU5+AU24 |
| Anger | 145 | A23, AU4, AU4+AU7, AU4+AU5, AU4+AU5+AU7,AU17+AU24, AU4+AU6+AU7, AU4+AU38 |
| Disgust | 17 | AU10, AU9, AU4+AU9, AU4+AU40, AU4+AU5+AU40, AU4+AU7+AU9, AU4+AU9+AU17, AU4+AU7+AU10,AU4+AU5+AU7+AU9, AU7+AU10 |
| Sadness | 34 | AU1,AU15,AU1+AU4,AU6+AU15,AU15+AU17 |
| Others | 102 | Others |

TABLE 4
AU labeling and frequency for database non-objective classification

| Classes | Number | Main AU labeling type |
|---|---|---|
| Positive | 44 | AUs needed for Happiness |
| Negative | 167 | AUs needed for disgust, anger,fear,sadness |
| Surprise | 37 | At least 1+2, 25 or 2 |
| Others | 125 | Others |

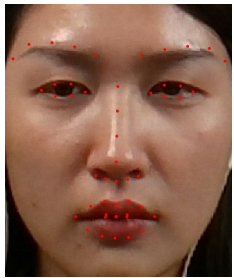

Fig.3. 49 landmark points on the face.

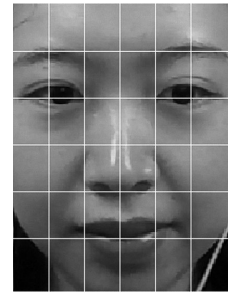

Fig.4. 6×6 block structure of a face frame.

The file "micro-expression.xls" records some information about the database sample to help you get the information quickly. The first column of the file records the participants' number, e. g. "1"; The second column is the sample for each participant, e. g. "1_1", representing the first sample of the first video material. Columns 3 to 5 are the onset frame, apex frame, and offset frame positions. The sixth column is the AUs of each sample. The seventh and eighth columns are the corresponding label categories of objective and non-objective classification, which help analyze and compare different categories.

# 4 DATABASE BASELINE EVALUATION

## 4.1 Preprocessing

### 4.1.1 Preprocessing of color sequences

Firstly, locate the faces. The incremental Parallel Cascade of Linear Regression (iPar-CLR) algorithm [56] was used to local the 49 face landmark points of expression sequence (see Fig.3).

Secondly, align the face. The inner eye corners of the first frame were calibrated to align the face using a non-reflective similarity transformation. Due to the short duration of the micro-expressions, slight head movements between sequence frames can be neglected. Therefore, the same transformation was used in the remaining frames of the sequence. Then, 49 facial landmark points of aligned sequences are relocated.

Finally, crop the face. The face crop proportions were determined using the calibrated inner eye corners and nasal spine of the first frame. The face scale was a 6×6 block size based on the horizontal between the inner eye corners and vertical distances between the nasal spine and the line connecting the inner eye corners [50] (see Fig.4). However, since the point clouds sequence and the color sequence were temporally and spatially aligned. To obtain the landmark points of point clouds, we saved the 49 landmark points for each frame of the color sequence before calibrated.

### 4.1.2 Preprocessing of point clouds sequences

Since the localization of 3D landmark points requires the assistance of 2D landmark points, we preprocessed 3D point clouds sequences after completing the preprocessing on the 2D image sequences.

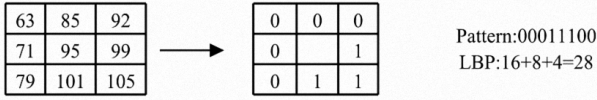

Fig.5. Eight-neighborhood LBP calculation process.

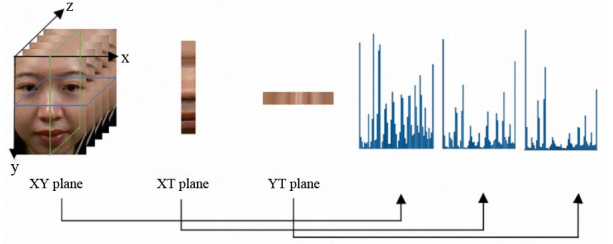

Fig.6. Textures and corresponding histograms of the three planes of micro-expressions.

First, the depth sequence frames of the micro-expressions were gradually smoothed, denoised, and hole-filled using spatial filtering, temporal filtering, and hole-filling operations. The depth images were then converted to point cloud data aligned with the corresponding color frames, and the face landmark points in 2D sequence frames corresponded to the point clouds frames to obtain 3D facial landmark points.

Second, since the nasal tip of a face is the lowest (or highest) point in the face point clouds [57], [58], [59], and locating the nasal tip by the lowest (or highest) point has been widely used in point clouds face extraction. The point clouds data was spherically cropped with a radius of 100 millimeters centered on the localized nasal tip landmark point to obtain the point clouds facial image.

Third, the facial point clouds of the remaining frames were aligned to the first frame using the ICP algorithm [60], [61], [62], and the landmark points were also aligned to the first frame using the same transformation.

## 4.2 Feature Extraction

### 4.2.1 LBP-TOP

The basic LBP operator compares the center pixel of an image with its neighbors. Take the eight-neighbor as an example. If the gray value of the center is greater than the gray value of the neighbor, the sign value is 0. Otherwise, the value is 1. The calculation process of LBP is shown in Fig.5.

For the center pixel $c$ and its $P$ neighboring pixels with the radius $R$, the LBP value of the center pixel is calculated as:

$$LBP_{P,R}(x_c, y_c) = \sum_{p=0}^{P-1} s(g_p - g_c) 2^p \quad (2)$$

where $(x_c, y_c)$ is the center pixel coordinate and $g_c$ is gray value. $g_p$ ($p = 0, 1, …, p$-1) is the $p$-th neighborhood gray value of the center pixel on a radius $R$. $2^p$ is the weight corresponding to the neighborhood pixel position. The function $s(x)$ is the sign function defined as:

$$s(x) = \begin{cases} 1 & x \geq 0; \\ 0 & x < 0. \end{cases} \quad (3)$$

For an input image, the statistical LBP histogram is used as the input image features.

To represent the dynamic change features of expressions, Zhao et al. [63] proposed the LBP-TOP algorithm as an extension of the LBP algorithm by adding LBP features in the XT, YT planes to represent the changes of expressions in the time dimension. Fig.6 shows XY, XT, YT plane texture and corresponding histogram features of the expression sequence. After counting all features of XY, XT, and YT planes, the three histograms are concatenated into a histogram vector as the final LBP-TOP feature vector. The parameters of each plane can be set as needed.

### 4.2.2 Curvature Feature---HK

Curvature is a characterization of the geometry of local surfaces. It is invariant to affine transformations like translation or rotation. Changes in the surface of the face reflect changes in facial expression, so it can be used as a 3D expression feature.

The principal curvature at a point on a surface is an eigenvalue that measures how the surface curves at that point in different directions with different magnitudes. The Gaussian curvature (K) is the product of the two principal curvatures of the surface, it measures the intrinsic curvature of the surface.

$$K = p_{min} p_{max} \quad (4)$$

where $p_{max}$ and $p_{min}$ are the maximum and minimum principal curvature of the surface. By determining the value of the Gaussian curvature, the curvature at the point on the surface can be analyzed, indicating whether the point is elliptical (K>0), parabolic (K=0), or hypersurface (K<0) [64].

The mean curvature (H) is the average of the principal curvatures, and it measures the degree of curvature of the surface in space.

$$H = \frac{p_{min} + p_{max}}{2} \quad (5)$$

Gaussian curvature and mean curvature reflect the convexity of the surface, but neither Gaussian nor mean curvature captures the local shape well [65]. The best way is to use the combination of K and H to represent different surface shapes. Table 5 shows the surface types and their geometrical descriptions under the nine combinations of K and H symbols [64], [66]. Where K> 0 and H= 0 are mathematically contradictory to each other and therefore do not exist, but it does not affect the analysis of the problem when it treats as a feature.

Any complex surface can be composed of these nine basic types. Gaussian and mean curvature can characterize the extrinsic shape, but it is clear that the features obtained by both curvatures are not as straightforward as using only one index. So, a feature that is sufficient to represent the same task is introduced - the shape index (SI) [65].

### 4.2.3 Curvature Feature---SI

The shape index is a quantitative measure of the shape of a surface and is calculated using the principal curvature of the surface spanned by the nearest neighbors of each vertex [67], [68]. The curvature values of all points on a 3D object can be mapped to [0,1].

TABLE 5
The local surface type of a point

| K | H | Surface type | Geometric description |
|---|---|---|---|
| >0 | >0 | Peak | The point is locally convex in all directions. |
| 0 | >0 | Ridge | The point is locally convex and flat in one direction. |
| <0 | >0 | Saddle ridge | Locally convex in most directions, locally concave in others. |
| >0 | 0 | / | / |
| 0 | 0 | Flat | Plane |
| <0 | 0 | Minimal surface | Locally convex and concave halves. |
| >0 | <0 | Pit | Locally concave in all directions. |
| 0 | <0 | Valley | Locally concave in all directions, flat in one main direction. |
| <0 | <0 | Saddle valley | Points are locally concave in most directions and locally convex in a small number of directions. |

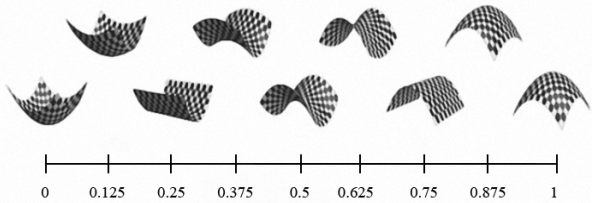

Fig.7. Shape index quantification [67].

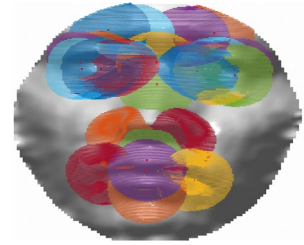

Fig.8. Local region of 23 landmark points on the face.

The shape index of each point is calculated by the maximum and minimum principal curvature of the surface, which needs to convert to local coordinates, and then calculates the eigenvalues of the Weingarten matrix by fitting the local surface with cubic polynomial. The shape index at point $p$ is:

$$SI_p = \frac{1}{2} - \frac{1}{\pi} \tan^{-1} \frac{p_{\max}(p) + p_{\min}(p)}{p_{\max}(p) - p_{\min}(p)} \quad (6)$$

Each different surface type corresponds to a unique shape index value.

In general, the shape index is scale quantized and transformed to nine quantization values that vary from concave to convex [65], quantified as shown in Fig. 7 [67,69].

In contrast, the SI features lack the flat feature of HK, but it provides a continuous change between salient shapes, so it can describe more subtle shape variations [69]. However, to analyze which surface representation feature could better represent expression changes, we discussed both forms of features for experimental analysis in the 3D feature extraction.

### 4.2.4 Feature representation and recognition method

The movement of micro-expressions is localized, the occurrence of expressions can be represented by the movement of landmark points and the changes of the local region of these points [34], [39], [70]. Facial micro-expressions are not manifested in all face regions, and different expressions occur in only a few of them. To remove unnecessary, we extracted the local features of the 23 land-marker points of the face based on the facial movements (see Fig.8).

To extract the motion information of facial micro-expressions more effectively, we selected a spherical domain of a certain radius around the point clouds landmark points to extract SI and HK features. Then quantized into nine components, and the quantized feature frequencies are used as 3D features to represent the facial spatial motion. The SI and HK features of the $j$-th landmark point in the $k$-th frame of the sequence can be represented as

$$SI_k^j = \left[ \frac{s_1}{M}, \frac{s_2}{M}, \frac{s_3}{M}, \frac{s_4}{M}, \frac{s_5}{M}, \frac{s_6}{M}, \frac{s_7}{M}, \frac{s_8}{M}, \frac{s_9}{M} \right] \quad (7)$$

$$HK_k^j = \left[ \frac{hk_1}{M}, \frac{hk_2}{M}, \frac{hk_3}{M}, \frac{hk_4}{M}, \frac{hk_5}{M}, \frac{hk_6}{M}, \frac{hk_7}{M}, \frac{hk_8}{M}, \frac{hk_9}{M} \right] \quad (8)$$

where $s_i$ and $hk_i$ denote the number of vertices of the corresponding SI and HK quantization scale, $M$ denotes the total local number of vertices. It could effectively represent the changes in the features corresponding to the landmark points in the sequence.

Inspired by [71], different micro-expressions have different motion regions, and the intensity of the motion also varies. We used the average intensity of the local circular domain of facial landmark points of the 2D mean difference image as weights to weight the local features of the 3D. Thus, the features corresponding to the $j$-th point in a sequence of length $m$ are:

$$SI_j = w_j \left( SI_1^j, SI_2^j, \cdots, SI_m^j \right) \quad (9)$$

$$HK_j = w_j \left( HK_1^j, HK_2^j, \cdots, HK_m^j \right) \quad (10)$$

Where $w_j$ is the weight of the feature of the $j$-th landmark point.

Finally, The SI and HK features of a sequence can be written as

$$SI = \left[ SI_1, SI_2, \cdots, SI_{23} \right] \quad (11)$$

$$HK = \left[ HK_1, HK_2, \cdots, HK_{23} \right] \quad (12)$$

Since the apex frame is the frame with the largest variation in the expression sequence, using the features of the onset and the apex frame are effective in representing the micro-expression changes while reducing time consumption.

Experimental analysis is performed using LIBSVM classifier, LOSO and 10-fold cross-validation methods. Since the random nature of the 10-fold cross-validation results, we take the average result of 10 times results as the recognition results in experiments.

We used a prediction probability weighting method to fuse 2D and 3D features, and prediction categories is based on the maximum value of the fusion probability. The weighted fusion method is

$$P = (1-a) \times p_1 + a \times p_2 \qquad (13)$$

where $p_1$ denotes the predicted probability of a 2D feature and $p_2$ means the predicted probability of a 3D feature, $a \in [0,1]$. Where 0 means that only 2D features are used and 1 means that only 3D features are used. The weight values are fixed for all samples in each classification.

### 4.3 Micro-Expression Recognition Evaluations

The recognition evaluation was performed on 373 processed sample of 33 subjects, and the experimental environment was Matlab 2021b.

We discussed the effect of different parameters on the recognition rate for 2D and 3D features under LOSO and 10-fold cross-validation for each classification. Further, we also evaluated the fusion results of other 2D algorithms and the 3D algorithm proposed in this paper. Accuracy and F-score were used to evaluate classification accuracy [32], [54]. F-score is a harmonic mean of the classification accuracy computed by precision and recall.

In non-objective classification experiments, the classification did not include samples from the sadness category because the number of sadness samples is too small to be accurately identified.

### 4.3.1 LBP-TOP parameter discussion

(1) Performance of LBP-TOP texture feature recognition under different radii

Table 6 analyzes the effect of LBP radii on the recognition rate in the XY, XT, and YT planes when the image is divided into $5 \times 5$ blocks, the overlap pixel is 0, and the neighborhoods are [8 8 8]. The x- and y-axis radii vary from 1 to 4, and the z-axis radius ranges from 2 to 4.

It can be seen that the radii affect the recognition rate. In the LOSO validation, there is a high recognition rate at radius $R_x = 1$, $R_y = 1$, $R_z = 4$. The 10-fold cross-validation has a high recognition rate at radius $R_x = 2$, $R_y = 2$, $R_z = 4$. These radii will be used as the basis in the discussion of other analyses below.

(2) LBP-TOP texture feature recognition performance under different overlapping pixels with different blocks

Block has a definite effect on the recognition rate. This is because micro-expressions occur locally, and a proper block can help to extract features effectively. To more fully analyze which blocks and overlaps have better recognition rate, in Tables 7, Table 8, Table 9 and 10, we discussed and analyzed 33 kinds of blocks from $3 \times 3$ to $8 \times 10$ and 7 kinds of overlaps such as 0, 5, 10, 15, 20, 25, and 30. Here, we only show the recognition results of some more commonly used blocks and some blocks with higher recognition accuracy in the experiment.

For objective classification, in LOSO, the highest recognition rate is 66.29% when the block structure is $3 \times 10$, and the overlap is 30. In 10-fold cross-validation, the higher recognition rate is 74.29% when the block is $7 \times 10$, and the overlap is 30.

Similarly, for non-objective classification, in the LOSO cross-validation experiment, the highest recognition rate is 57.16% when the block is set to $5 \times 5$ and overlap is 5. In 10-fold cross-validation, the highest recognition rate is 70.33% when the block is set to $8 \times 8$ and the overlap is 15. Of course, this is only a partial test and does not indicate the absolute best results. Since the fusion performance of different blocks may also be different, in the fusion of 2D and 3D feature results, we performed fusion experiments on several blocks with relatively high results to analyze the fusion performance.

TABLE 6
Recognition rate for LBP-TOP feature at different radii (%)

| $R_x$ $R_y$ $R_t$ | Objective classification | | Non-objective classification | |
|---|---|---|---|---|
| | LOSO | 10-fold | LOSO | 10-fold |
| 1  1  2 | 50.02 | 70.50 | 48.86 | 67.06 |
| 1  1  3 | 56.76 | 71.42 | 50.07 | 68.50 |
| 1  1  4 | **59.66** | 72.60 | **53.19** | 69.52 |
| 2  2  2 | 48.77 | 71.10 | 45.46 | 67.09 |
| 2  2  3 | 54.40 | 71.60 | 48.02 | 68.61 |
| 2  2  4 | 56.10 | **73.58** | 49.74 | **69.62** |
| 3  3  2 | 45.44 | 69.99 | 45.14 | 64.83 |
| 3  3  3 | 51.28 | 71.83 | 46.40 | 67.01 |
| 3  3  4 | 53.21 | 72.70 | 47.76 | 67.86 |
| 4  4  2 | 46.26 | 70.81 | 42.00 | 65.23 |
| 4  4  3 | 51.55 | 71.79 | 44.30 | 66.96 |
| 4  4  4 | 50.80 | 71.25 | 45.85 | 67.33 |

Table 7
Recognition rate for LBP-TOP feature using LOSO validation for objective classification (%)

| Blocks | Overlaps | | | | | | |
|---|---|---|---|---|---|---|---|
| | 0 | 5 | 10 | 15 | 20 | 25 | 30 |
| 5*5 | 59.66 | 61.33 | 62.37 | 62.28 | 63.01 | 63.71 | 63.80 |
| 6*6 | 57.74 | 62.46 | 62.75 | 61.51 | 60.26 | 63.53 | 64.10 |
| 8*8 | 59.53 | 63.65 | 62.15 | 60.83 | 62.29 | 60.42 | 63.33 |
| 3*6 | 63.92 | 63.41 | 66.17 | 62.72 | 65.31 | 65.29 | 64.52 |
| 3*10 | 61.49 | 58.10 | 64.04 | 65.64 | 66.01 | 64.74 | 66.29 |
| 7*9 | 60.22 | 61.76 | 60.58 | 58.05 | 62.00 | 64.88 | 65.76 |

Table 8
Recognition rate for LBP-TOP feature using 10-fold cross validation for objective classification (%)

| Blocks | overlaps | | | | | | |
|---|---|---|---|---|---|---|---|
| | 0 | 5 | 10 | 15 | 20 | 25 | 30 |
| 5*5 | 73.58 | 72.28 | 72.60 | 71.88 | 72.89 | 72.07 | 71.67 |
| 6*6 | 73.16 | 73.41 | 73.26 | 73.12 | 73.37 | 72.61 | 73.84 |
| 8*8 | 72.85 | 71.99 | 73.23 | 74.12 | 73.19 | 73.96 | 72.97 |
| 3*6 | 73.37 | 73.16 | 73.31 | 73.66 | 73.66 | 73.15 | 73.67 |
| 4*8 | 73.50 | 72.84 | 73.82 | 74.12 | 73.98 | 73.41 | 73.03 |
| 7*10 | 73.83 | 71.08 | 74.20 | 72.54 | 73.83 | 74.16 | 74.29 |

Table 9
Recognition rate for LBP-TOP feature using LOSO validation for non-objective classification (%)

| Blocks | Overlaps | | | | | | |
|---|---|---|---|---|---|---|---|
| | 0 | 5 | 10 | 15 | 20 | 25 | 30 |
| 4*4 | 49.05 | 51.02 | 52.03 | 52.23 | 53.07 | 55.53 | 52.74 |
| 5*5 | 53.19 | 57.16 | 56.31 | 57.11 | 53.32 | 53.60 | 52.49 |
| 6*6 | 53.49 | 53.21 | 50.18 | 50.66 | 51.99 | 48.82 | 54.80 |
| 8*8 | 54.58 | 54.16 | 52.92 | 51.40 | 50.39 | 49.57 | 52.77 |
| 3*4 | 54.46 | 56.42 | 54.51 | 55.79 | 53.02 | 51.80 | 50.96 |
| 4*9 | 51.82 | 52.10 | 48.21 | 51.47 | 50.07 | 49.06 | 53.66 |

Table 10
Recognition rate for LBP-TOP feature using 10-fold cross validation non-objective classification (%)

| Blocks | overlaps | | | | | | |
|---|---|---|---|---|---|---|---|
| | 0 | 5 | 10 | 15 | 20 | 25 | 30 |
| 5*5 | 69.62 | 68.99 | 68.50 | 67.64 | 66.11 | 66.42 | 66.79 |
| 6*6 | 68.68 | 68.58 | 66.72 | 68.07 | 66.35 | 65.71 | 67.09 |
| 8*8 | 68.85 | 69.12 | 68.07 | 70.33 | 68.44 | 67.82 | 67.27 |
| 5*8 | 69.04 | 68.42 | 68.31 | 69.55 | 68.97 | 68.41 | 66.96 |
| 5*9 | 69.29 | 68.22 | 67.96 | 67.26 | 67.88 | 68.04 | 67.44 |
| 6*10 | 68.58 | 68.06 | 68.68 | 65.85 | 68.72 | 66.36 | 67.90 |

### 4.3.2 Recognition performance in fusion mode
Table 11 shows the results of the fusion features under objective classification and non-objective classification. Since different 3D neighborhood radii features have different 3D recognition performance, we also conducted experimental analysis of 3D features with different radii, and analyzed fusion results of 2D features and 3D features with different radii. Only the optimal fusion results are shown here.

For objective classification, in LOSO validation, it has the better fusion results when setting the block as 7×9 and the overlap as 30. The highest fusion recognition rate of LBP-TOP and SI features is 67.97%, which is 2.21% higher than that of LBP-TOP. When fused with HK features, the highest recognition rate is 67.34%, which is 1.58% higher than that of LBP-TOP. The highest recognition rate is 68.73%, which is 2.97% higher than SI and HK. F1-score is 0.4311. In 10-fold cross-validation, the fusion result with 7×10 block and 30 overlap was taken as the baseline result. The fusion recognition rate of LBP-TOP and SI features was up to 76.39%, an improvement of 2.1% compared to LBP-TOP. When fused with HK features, the highest recognition rate is 75.93%, which is 1.64% higher than that of LBP-TOP. When fused with SI+HK features, the recognition rate is 76.17%, which is 1.88% higher. But in the above results, F1-score did not improve significantly.

Table 11
Fusion results for LBP-TOP feature and 3D features

| Features | Objective | | | | Non-objective | | | |
| | LOSO | | 10-fold | | LOSO | | 10-fold | |
| | Acc(%) | F1-score | Acc(%) | F1-score | Acc(%) | F1-score | Acc(%) | F1-score |
|---|---|---|---|---|---|---|---|---|
| LBP-TOP(2D) | 65.76 | 0.4227 | 74.29 | 0.6358 | 56.42 | 0.5136 | 69.29 | 0.6776 |
| SI(3D) | 53.31 | 0.2784 | 61.37 | 0.3464 | 41.43 | 0.2888 | 51.23 | 0.3995 |
| HK(3D) | 51.39 | 0.2485 | 59.52 | 0.3199 | 44.73 | 0.2343 | 49.91 | 0.3901 |
| SI+HK(3D) | 52.61 | 0.2561 | 60.36 | 0.3444 | 42.42 | 0.3257 | 52.83 | 0.4607 |
| LBP-TOP(2D)+SI(3D) | 67.98 | **0.4311** | **76.39** | 0.6355 | 58.55 | 0.5285 | 71.39 | 0.6770 |
| LBP-TOP(2D)+HK(3D) | 67.34 | 0.4290 | 75.93 | 0.6206 | 58.10 | **0.5312** | **71.72** | **0.6874** |
| LBP-TOP(2D)+SI+HK(3D) | **68.73** | 0.4269 | 76.17 | 0.6300 | **58.67** | 0.5302 | 71.51 | 0.6845 |

Table 12
Best identification results for DSME database using
deep learning algorithms and 3D algorithm

| Features | Acc(%) | F1-score |
|---|---|---|
| 2D(3DSTCNN[72]) | 72.48 | 0.4524 |
| 2D(3DSTCNN)+3D | 74.56 | 0.5242 |
| 2D(EMR+AT[73]) | 70.64 | 0.4959 |
| 2D(EMR+AT)+3D | 75.32 | 0.5106 |
| 2D(STSTNET[74]) | 73.56 | 0.6136 |
| 2D(STSTNET)+3D | 76.50 | 0.6176 |

For non-objective classification, in LOSO validation, it has the better fusion results when setting the block as 3×4 and the overlap as 5. The highest fusion recognition rate of LBP-TOP and SI features is 58.55%, which is 2.13% higher than that of LBP-TOP. When fused with HK features, the highest recognition rate is 58.10%, which is 1.68% higher than that of LBP-TOP. The highest recognition rate is 58.67 %, which is 2.25% higher than SI and HK. F1-score is 0.5312. In 10-fold cross-validation, the fusion result with 5×9 block and 0 overlap was taken as the baseline result. The fusion recognition rate of LBP-TOP and SI features was up to 71.39%, an improvement of 2.1% compared to LBP-TOP. When fused with HK features, the highest recognition rate is 71.72% which is 2.43% higher than that of LBP-TOP. When fused with SI+HK features, the recognition rate is 71.51%, which is 2.22% higher. For F1 up to 0.6874, but the LOSO result is lower than 10-fold.

The above results fully demonstrate that the combination of 3D features can improve the recognition rate of micro-expressions, both for objective and non-objective classification. It also proves that the 3D spatiotemporal information is useful for representing micro-expressions. Further, we analyzed the prediction results in 3D and 2D and found that we could obtain a great improvement in 2D to identify samples that were not accurately identified in 2D, but the recognition rate after fusing 3D in this paper did not reach this desired result and sometimes the fusion did not have a significant effect, which may have some relationship with the processing of 3D data and the fusion method of features.

### 4.3.3 Others methods evaluations

We used other state-of-the-art 2D feature extraction algorithms and fused them with the 3D features extracted in this paper to further illustrate the performance of the database. The algorithm of [72], [73],[74] are the deep neural network algorithms. The recognition results of these algorithms on our database are shown in Table 12. Because the sample size is small, referring to the existed research methods of micro-expression deep learning, we reclassify the samples into three categories according to AU: positive, negative, surprise, and discard the samples of 'Others' class, and use the LOSO validation method for database estimation. It can be seen that the three algorithms compared with the results after fusion with 3D, 3DSTCNN improved by 2.08%, STSTNET improved by 4.68%, and EMR+AT improved by 2.94%. 3DSTCNN is a spatio-temporal 3DCNN, and the recognition rate is a bit lower than the other two. EMR+AT has a relatively high recognition rate of 2D features because it amplifies the motion and uses macro-expressions. And the F1-score have improved, so the classification results of the database are all improved somewhat with the assistance of 3D features. For the fusion results, the performance of deep learning algorithms is slightly better due to fewer categories, but further research is needed to come up with more efficient algorithms.

## 5 DISCUSSION

The contribution of this paper is to create a new micro-expressions database and to verify that combining 2D texture features with 3D spatial features of micro-expression changes can improve the inadequacy of relying only on 2D information to express micro-expressions. However, there are still some limitations in the database.

Two types of encoding. There is no one unified way of encoding micro-expressions. It is impossible to determine which way is appropriate and accurate. Since different people express emotions in different ways, it may not be comprehensive to classify emotions only by facial expressions. Other internal differences and environmental factors should also be considered. Therefore, facial micro-expressions were classified by objective coding method based on FACS, and emotions were classified by non-objective method combined with video content and subjective self-reports.

The size of the database is limited. Although the database offers the possibility to study micro-expressions in more depth, the size of the database may be insufficient due to the difficulty of eliciting micro-expressions and the extremely time-consuming manual coding, similar to the currently available databases. Currently, we are still extending the database. In addition, the number of categories is unbalanced among categories which will greatly affect the results of estimations. We discarded meaningless facial movements, but since some expressions occur with a blinking motion, these expressions cannot be excluded due to blinking and retained in this database. Of course, this also affects the recognition performance to some extent.

Unlike macro-expressions, 3D micro-expressions are still at the stage of research and exploration. There is no or little research related to 3D micro-expressions, the 3D feature extraction algorithm proposed in this paper is still in the initial stage. For algorithms using deep learning, this paper only explores the fusion of three outstanding deep learning algorithms on the existing 2D micro-expression database with the 3D features in this paper, but it does not show the effectiveness of such algorithms for this database. The 3D information in the DSME-3D database may have some deficiencies in data processing as well as feature extraction, so the recognition performance is not ideal, so further research is needed. The DSME-3D database will be available for download with permission. The database contains cropped 2D and 3D micro-expression sequences and uncropped micro-expression sequences.

## 6 CONCLUSION

This paper constructs a new spontaneous dynamic micro-expression database, which is the first micro-expression database containing dynamic 3D spatial data. It aims to improve recognition performance by incorporating spatio-temporal deformation features of micro-expressions for more detailed and subtle facial behaviors. This database was created in a controlled laboratory environment, where high-speed infrared binocular cameras simultaneously record dynamic 2D and 3D information of expressions. The database includes 373 samples, each containing a 2D texture sequence and a corresponding 3D point cloud sequence, marking the onset, apex, and offset frames of the expression sequence, coding and classifying the samples using both objective and non-objective classification. It is useful to study the validity of the classification method. Evaluation results of 2D, 3D features, and classification fusion using LOSO and 10-fold cross-validation methods were presented for comparative analysis in future studies. Other 2D features are also fused with 3D features to perform database performance analysis.

The DSME-3D database makes up for the lack of 3D spatial information in previous micro-expression databases by capturing the dynamic 3D information of micro-expressions. However, due to the poor robustness of the extracted 3D features, and the advantages of 3D information in expression recognition cannot be fully utilized. In the future, 3D data processing and feature representation will continue to be the key to our research, and more powerful methods are needed to make data processing and visualization faster and more accurate. Simultaneously, the 2D feature extraction algorithm for facial micro-expressions and the adaptive fusion algorithm of 2D and 3D features will be our next research direction.

## ACKNOWLEDGMENTS

The authors would like to thank the qualified coders involved in the FACS coding of the database, all those who participated in the emotion elicitation experiments, and all those who have helped in the creation of the database.

## REFERENCES

[1] P. Ekman, and W. V. Friesen, "Nonverbal leakage and clues to deception," *Psychiatry*, vol. 32, no. 1, pp. 88–106, 1969.

[2] W. -J. Yan, Q. Wu, J. Liang, Y. -H. Chen, and X. Fu, "How Fast are the leaked facial expressions: The duration of micro-expressions," *J. Nonverbal Behav.*, vol. 37, no. 4, pp. 217-230, 2013.

[3] W.-J. Yan, S.-J. Wang, Y.-J. Liu, Q. Wu, and X. Fu, "For micro-expression recognition: Database and suggestions," *Neurocomputing*, vol. 136, pp. 82-87, 2014.

[4] Z. Xia, X. Feng, J. Peng, X. Peng, and G. Zhao, "Spontaneous micro-expression spotting via geometric deformation modeling," *Comput. Vision and Image Understanding*, vol. 147, pp. 87-94, 2016.

[5] X. Li, X. Hong, A. Moilanen, X. Huang, T. Pfister, G. Zhao, and M. Pietikinen, "Towards reading hidden emotions: A comparative study of spontaneous micro-expression spotting and recognition methods," *IEEE Trans. Affect. Comput.*, 2015, vol. 9, no. 4, pp. 563-571.

[6] D. Matsumoto, and H. S. Hwang, "Evidence for training the ability to read microexpressions of emotion," *Motivation and Emotion*, vol. 35, no. 2, pp. 181-191, 2011.

[7] M. Takalkar, M. Xu, Q. Wu, and Z. Chaczko, "A survey: facial micro-expression recognition," *Multimedia Tools and Appl.*, vol. 77, no. 15, pp. 19301-19325, 2018.

[8] O. Yee-Hui, S. John, L. N. A. Cat, P. R. C.-W., and V. M. J. F. i. P. Baskaran, "A survey of automatic facial micro-expression analysis: Databases, methods and challenges," *Comput. Vision and Pattern Recognit.*, vol. 9, pp. 1128-, 2018.

[9] Gupta, Puneet, "MERASTC: Micro-expression Recognition using

Effective Feature Encodings and 2D Convolutional Neural network." *IEEE Trans. Affect. Comput.*, pp (99):1-1,2021

[10] J. H. Yu, C. Y. Zhang, Y. Song, and W. D. Cai, "ICE-GAN: Identity-aware and Capsule-Enhanced GAN for Micro-Expression Recognition and Synthesis." (2020). *arXiv preprint arXiv: 2005.04370.*

[11] L. Lei, T. Chen, S. Li, and J. Li, "Micro-expression Recognition Based on Facial Graph Representation Learning and Facial Action Unit Fusion," *in Proc. 2021 IEEE/CVF Conf. Comput. Vision and Pattern Recog. Workshops (CVPRW)*, 2021, pp. 1571-1580, doi: 10.1109/CVPRW53098.2021.00173.

[12] J. See, M. H. Yap, J. T. Li, X. P. Hong, and S.-J. Wang, "MEGC 2019 – The Second Facial Micro-Expressions Grand Challenge." *in Proc. IEEE Conf. Automat. Face and Gesture Recog.*, 2019.

[13] L.-J. Yin, X.-Z. Wei, Y. S, J. Wang, and M. J. Rosato, "A 3D facial expression database for facial behavior research." *in Proc. 7th IEEE Conf. Automat. Face and Gesture Recog. (FGR06)*, 2006, pp. 211-216.

[14] M. Behzad, N. Vo, X. B. Li, and G.Y. Zhao, "Sparsity-Aware Deep Learning for Automatic 4D Facial Expression Recognition." (2020). *arXiv:2002.03157*

[15] P. Zarbakhsh, and H. Demirel, "4D facial expression recognition using multimodal time series analysis of geometric landmark-based deformations," *The Visual Comput.*, vol. 36, no. 5, pp. 951-965, 2020.

[16] R. Weber, J. Li, C. Soladié, and R. Séguier, "A Survey on Databases of Facial Macro-expression and Micro-expression," *in Proc. Int. Joint Conf. Comput. Vision, Imag. and Comput. Graph. Theory and Appl.*, 2019, pp. 298-325.

[17] P. Lucey, J. F. Cohn, T. Kanade, J. Saragih, Z. Ambadar, and I. Matthews, "The extended cohn-kanade dataset (ck+): A complete dataset for action unit and emotion-specified expression," *in Proc. 2010 IEEE Comput. Soc. Conf. Comput. Vision and Pattern Recog. Workshops (CVPRW)*. IEEE, 2010, pp. 94–101.

[18] S. M. Mavadati, M. H. Mahoor, K. Bartlett, P. Trinh, and J. F. Cohn, "DISFA: A Spontaneous Facial Action Intensity Database," *IEEE Trans. Affect. Comput.*, vol. 4, no. 2, pp. 151-160, 2013.

[19] D. McDuff, R. El Kaliouby, T. Senechal, M. Amr, J. F. Cohn, and R. Picard, "Affectiva-mit facial expression dataset (am-fed): Naturalistic and spontaneous facial expressions collected" in the wild"," *in Proc. 2013 IEEE Conf. Comput. Vision and Pattern Recog. Workshops (CVPRW)*. IEEE, 2013, pp. 881–888.

[20] S. Wang, Z. Liu, Z. Wang, G. Wu, P. Shen, S. He, and X. Wang, "Analyses of a multimodal spontaneous facial expression database," *IEEE Trans. Affect. Comput.*, vol. 4, no. 1, pp. 34-46, 2013.

[21] X. Zhang, L. Yin, J. F. Cohn, S. Canavan, M. Reale, A. Horowitz, P. Liu, and J. M. Girard, "BP4D-Spontaneous: a high-resolution spontaneous 3D dynamic facial expression database," *Imag. and Vision Comput.*, vol. 32, no. 10, pp. 692-706, 2014.

[22] X. Zhang, L.-J. Yin, J.F. Cohn, S. Canavan, M. Reale, A. Horowitz, and P. Liu, "A high resolution spontaneous 3D dynamic facial expression database", *in Proc. 10th IEEE Int. Conf. Automat. Face and Gesture Recog.*, 2013.

[23] Z. Zhang, J.M. Girard, Y. Wu, X. Zhang, P. Liu, U. Ciftci, S. Canavan, M. Reale, A. Horowitz, H.Y.Yang, J.F. Cohn, J. Qiang, and L.J.Yin, "Multimodal spontaneous emotion corpus for human behavior analysis" *in Proc. IEEE Conf. Comput. Vision and Pattern Recog.* 2016, pp. 3438-3446.

[24] M. Shreve, S. Godavarthy, D. Goldgof, and S. Sarkar, "Macro-
and micro-expression spotting in long videos using spatio-temporal strain." *in Proc. 2011 IEEE Conf. Automat. Face and Gesture Recog.*, 2011, pp. 51-56.

[25] S. Polikovsky, Y. Kameda, and Y. Ohta, "Facial micro-expressions recognition using high speed camera and 3D-gradient descriptor." *in Proc. 3rd Int. Conf. Crime Detection and Prevention (ICDP 2009)*. IET, 2009, pp. 1-6.

[26] X. Li, T. Pfister, X. Huang, G. Zhao, and M. Pietikainen, "A spontaneous micro-expression database: Inducement, collection and baseline," *in Proc. 10th IEEE Int. Conf. and Workshops on Automat. Face and Gesture Recog. (FG)*. IEEE, 2013, pp. 1–6.

[27] W.-J. Yan, Q. Wu, Y.-J. Liu, S.-J. Wang, and X. FU, "CASME Database: A dataset of spontaneous micro-expressions collected from neutralized faces," *in Proc. 10th IEEE Conf. Automat. Face and Gesture Recog.*, 2013, pp. 1–7.

[28] W.-J. Yan, X. Li, S.-J. Wang, G. Zhao, Y.-J. Liu, Y.-H. Chen, and X. Fu, "CASME II: An improved spontaneous micro-Expression database and the baseline evaluation," *PLoS ONE*, vol. 9, no. 1, pp. e86041, 2014.

[29] F. Qu, S. Wang, W. Yan, H. Li, S. Wu, and X. Fu, "CAS(ME)²: A database for spontaneous macro-expression and micro-expression spotting and recognition," *IEEE Trans. Affect. Comput.*, vol. 9, no. 4, pp. 424-436, 2018.

[30] A. K. Davison, C. Lansley, N. Costen, K. Tan, and M. H. Yap, "SAMM: A spontaneous micro-facial movement dataset," *IEEE Trans. Affect. Comput.*, vol. 9, no. 1, pp. 116-129, 2018.

[31] X.Y. Ben, Y.Ben, J.P. Zhang, S.-J Wang, K. Kpalma, W.M.Meng, and Y.J.Liu, "Video-based Facial Micro-Expression Analysis: A Survey of Datasets, Features and Algorithms," *in IEEE Trans. Pattern. Anal. and Mach. Intel.*, doi: 10.1109/TPAMI.2021.3067464.

[32] W. Merghani, A. K. Davison, and M. H. Yap, "A review on facial micro-expressions analysis: Datasets, features and metrics," *arXiv preprint arXiv:1805.02397*, 2018.

[33] Y. Wang, J. See, R. C. W. Phan, and Y.-H. Oh, "LBP with six intersection points: Reducing redundant information in LBP-TOP for micro-expression recognition." *in Proc. Asian Conf. Comput. Vision -ACCV 2014*, Springer, 2014, pp. 525-537.

[34] Y. Zong, X. Huang, W. Zheng, Z. Cui, and G. Zhao, "Learning from hierarchical spatiotemporal descriptors for micro-expression recognition," *IEEE Trans. Multimedia*, vol. 20, no. 11, pp. 3160-3172, 2018.

[35] X. Huang, S. Wang, G. Zhao, and M. Piteikäinen, "Facial micro-expression recognition using spatiotemporal local binary pattern with integral projection." *in Proc. 2015 IEEE Int. Conf. Comput. Vision Worshop (ICCVW)*., 2015, pp. 1-9.

[36] X. Huang, S.-J. Wang, X. Liu, G.-Y Zhao, X.-Y. Feng, and M. Pietikäinen, "Discriminative spatiotemporal local binary pattern with revisited integral projection for spontaneous facial micro-expression recognition," *IEEE Trans. Affect. Comput.*, vol. 10, no. 1, pp. 32-47, 2019.

[37] M. Yu, Z. Q. Guo, Y. Yu, Y. Wang, and S. -X. Cen, "Spatiotemporal feature descriptor for micro-expression recognition using local cube binary pattern," *IEEE Access*, vol. 7, pp. 1-1, 2019.

[38] C. Guo, J. Liang, G. Zhan, Z. Liu, M. Pietikäinen, and L. Liu, "Extended local binary patterns for efficient and robust spontaneous facial micro-expression recognition," *IEEE Access*, vol. 7, pp. 174517-174530, 2019.

[39] Y. Liu, J. Zhang, W. Yan, S. Wang, G. Zhao, and X. Fu, "A Main Directional Mean Optical Flow Feature for Spontaneous Micro-Expression Recognition," *IEEE Trans. Affect. Comput.*, vol. 7, no.

4, pp. 299-310, 2016.

[40] Y J Liu, B J Li, Y K Lai. "Sparse MDMO: Learning a Discriminative Feature for Spontaneous Micro-Expression Recognition," *IEEE Trans. Affect. Comput.*, vol. 12, no. 1, pp.1-1,2018.

[41] S. T. Liong, J. See, C. W. Phan, A. C. L. Ngo, Y. H. Oh, and K. S. Wong, "Subtle expression recognition using optical strain weighted features", *in Proc. Asian Conf. Comput. Vision -ACCV 2014 Workshops*, 2014, pp. 1-14.

[42] S.-T. Liong, J. See, R. C.-W. Phan, and K. J. A. Wong, "Less is More: Micro-expression Recognition from Video using Apex Frame," *ArXiv*, vol. abs/1606.01721, 2018.

[43] S.-T. Liong, Y. S. Gan, W. -C. Yau, Y. -C Huang, and T. L. Ken, "OFF-ApexNet on Micro-expression Recognition System," *ArXiv*:1805.08699, 2018.

[44] M. Verma, S. K. Vipparthi, G. Singh, and S. Murala, "LEARNet: Dynamic imaging network for micro expression recognition," *IEEE Trans. Image Process.*, vol. 29, pp. 1618-1627, 2020.

[45] Z. Xia, X. Hong, X. Gao, X. Feng, and G. Zhao, "Spatiotemporal recurrent convolutional networks for recognizing spontaneous micro-expressions," *IEEE Trans. Multimedia*, vol. 22, no. 3, pp. 626-640, 2020.

[46] S. R. Zhao, H. Q. Tao, Y. S. Zhang, T. Xu, K. Zhang, Z. K. Hao, and E. H. Chen, "A two-stage 3D CNN based learning method for spontaneous micro-expression recognition." *Neurocomputing*, pp, 276-289, 2021.

[47] X. H Zhao, H. M.Ma, and R.Q Wang, "STA-GCN: Spatio-Temporal AU Graph Convolution Network for Facial Micro-expression Recognition," *Pattern. Recog. and Comput. Vision. PRCV 2021. Lecture Notes in Computer Science*, vol 13019. Springer, Cham. https://doi.org/10.1007/978-3-030-88004-0_7.

[48] S.-J. Wang, H.-L. Chen, W.-J. Yan, Y.-H. Chen, and X. Fu, "Face recognition and micro-expression recognition based on discriminant tensor subspace analysis plus extreme learning machine," *Neural Process. Lett.*, vol. 39, no. 1, pp. 25-43, 2014.

[49] S.-J. Wang, W.-J. Yan, T. Sun, G. Zhao, and X. Fu, "Sparse tensor canonical correlation analysis for micro-expression recognition," *Neurocomputing*, vol. 214, pp. 218-232, 2016.

[50] Z. Lu, Z. Luo, H. Zheng, J. Chen, and W. Li, "A delaunay-based temporal coding model for micro-expression recognition." *in Comput. Vision - ACCV 2014 Workshops*, Springer, 2015, pp. 698-711.

[51] J. J. Gross, and R. W. Levenson, "Emotion elicitation using films." *Cognition & Emotion.* vol. 9, no.1, pp.9-28, 2007.

[52] Y. L. Deng, M. Yang, and R. L. Zhou. "A New Standardized Emotional Film Database for Asian Culture." *Frontiers in Psychology*, 2017, vol. 8, 1941.

[53] P. Ekman, W. V. Friesen, and J. C. Hager, *Facial Action Coding System: Investigator's Guide*. Salt Lake: Network Information Research Corporation, 2002.

[54] A. K. Davison, W. Merghani, and M. H. Yap, "Objective classes for micro-facial expression recognition," *J. Imag.*, vol. 4, no. 10, 2017.

[55] P. Ekman, W. V. Friesen, and J. C. Hager, *Facial Action Coding System: The Manual on CD Rom.* Salt Lake: Network Information Research Corporation, 2002.

[56] A. Asthana, S. Zafeiriou, S. Cheng, and M. Pantic, "Incremental face alignment in the wild." *in Proc. 2014 IEEE Conf. Comput. Vision and Pattern Recog.*, 2014, pp. 1859-1866.

[57] V. Le, H. Tang, and T. S. Huang, "Expression recognition from 3D dynamic faces using robust spatio-temporal shape features." *in Face and Gesture 2011*, 2011, pp. 414-421.

[58] S. Berretti, A. del Bimbo, and P. Pala, "Automatic facial expression recognition in real-time from dynamic sequences of 3D face scans," *The Visual Comput.*, vol. 29, no. 12, pp. 1333-1350, 2013.

[59] R. Sala Llonch, E. Kokiopoulou, I. Tošić, and P. Frossard, "3D face recognition with sparse spherical representations," *Pattern Recog.*, vol. 43, no. 3, pp. 824-834, 2010.

[60] X. Li, Q. Ruan, G. An, and Y. Jin, "Automatic 3D facial expression recognition based on polytypic Local Binary Pattern." *in Proc. 12th Int. Conf. Signal Process. (ICSP)*, 2014, pp. 1030-1035.

[61] S. P. S., S. P., S. Tripathi, and L. R., "Emotion recognition from facial expressions of 4D videos using curves and surface normals," *Intell. Human Comput. Interaction. IHCI 2016*, Springer, 2016, pp. 51-64.

[62] X. Li, Q. Ruan, Y. Jin, G. An, and R. Zhao, "Fully automatic 3D facial expression recognition using polytypic multi-block local binary patterns," *Signal Process.*, vol. 108, pp. 297-308, 2015.

[63] G. Zhao, and M. Pietikainen, "Dynamic texture recognition using local binary patterns with an application to facial expressions," *IEEE Trans. Pattern Anal. Mach. Intell.*, vol. 29, no. 6, pp. 915-928, 2007.

[64] S.-L. Liu, R.-R. Zhou, and L.-L. An, "Data segmentation algorithm for a triangular mesh model." *J. Nanjing Univ. of Aeronaut. & Astronaut.*, vol.35, no. 6, pp. 77-82, 2003.

[65] J. J. Koenderink, and A. J. van Doorn, "Surface shape and curvature scales," *Image and Vision Comput.*, vol. 10, no. 8, pp. 557-564, 1992.

[66] Y.-M. Wang, C.-H. Wu, G. Pan, "New method for facial feature detection based on range data." *J. Zhejiang Univ. (Eng. Sci.)*, vol. 39, no. 5, pp. 652-656, 2005.

[67] Canavan, S., Y. Sun, X. Zhang, and L.-J. Yin, "A dynamic curvature based approach for facial activity analysis in 3D space." *in Proc. 2012 IEEE Comput. Soc. Conf. Vision and Pattern Recog. Workshops(CVPRW)*, IEEE, 2012, pp. 14-19.

[68] C. Sindhuja, and K. Mala, "Landmark identification in 3D image for facial expression recognition." *in Proc. 2015 Int. Conf. Comput. & Commun. Technologies (ICCCT)*, 2015, pp. 338-343.

[69] Dorai, Chitra, Jain, et al. "COSMOS--a representation scheme for 3D free-form objects". *IEEE Trans. Pattern Anal. & Mach. Intell.*, vol. 9, no.10, pp. 1115-1130, 1997.

[70] H. Li, H. Ding, D. Huang, Y. Wang, X. Zhao, J.-M. Morvan, and L. Chen, "An efficient multimodal 2D + 3D feature-based approach to automatic facial expression recognition," *Comput. Vision and Image Understanding*, vol. 140, pp. 83-92, 2015.

[71] A. Maalej, B. Ben Amor, M. Daoudi, A. Srivastava, and S. Berretti, "Local 3D shape analysis for facial expression recognition." *in Proc. 20th Int. Conf. Pattern Recog.*, 2010, pp. 4129-4132.

[72] Reddy S, Karri S T, Dubey S R and Mukherjee S. "Spontaneous Facial Micro-Expression Recognition using 3D Spatiotemporal Convolutional Neural Networks," *in Proc. 2019 Int. Joint Conf. Neural Networks (IJCNN)*, 2019.

[73] Y. Liu, H. Du, L. Zheng, and T. Gedeon, "A neural microexpression recognizer," *in Proc. IEEE Int. conf. automatic face & gesture recognition (FG)*. IEEE, 2019, pp. 1–4.

[74] S.T. Liong, Y. S. Gan, J. See, H. Q. Khor and Y. C. Huang, "Shallow Triple Stream Three-dimensional CNN (STSTNet) for Micro-expression Recognition," *in Proc. IEEE Int. conf. Automatic Face & Gesture Recognition (FG).* IEEE, 2019, pp. 1-5.

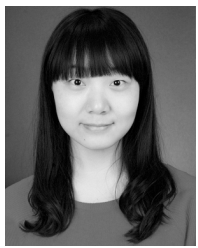

**Fengping Wang** received the M.S. degree from Xi'an University of Technology, Xi'an, China, in 2016, and she is currently pursuing the Doctoral degree with the School of Information and Communications Engineering, Xi'an Jiaotong University. Her current research interests include face detection and facial micro-expression recognition.

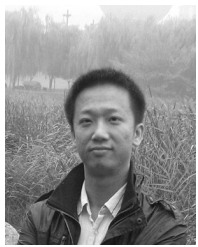

**Jie Li** received the Ph.D. degree in Electronics Science and Technology from Xi'an Jiaotong University in 2012, and was discharged from the Information and Communication Engineering Postdoctoral Station in 2014. He is currently a Professor and Ph.D. supervisor. He is currently research on spectral and polarized image acquisition and processing techniques.

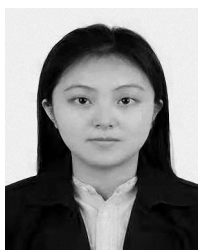

**Siqi Zhang** received the M.S. degree from University of Birmingham, Birmingham, United Kingdom, in 2015. Her current research interests include electromagnetic sensor networks.

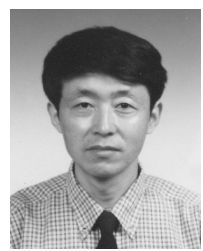

**Chun Qi** received the Ph.D. degree from Xi'an Jiaotong University, Xi'an, China, in 2000. He is currently a Professor and Ph.D. supervisor with the School of Information and Communications Engineering, Xi'an Jiaotong University. His current research interests mainly include image enhancement, detection and tracking, compression coding, spectral imaging and processing.

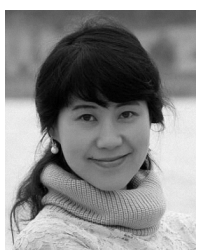

**Yun Zhang** received the Ph.D. degree in Northwestern Polytechnical University, and then worked in Hong Kong Polytechnic University for two years of eye tracking related research. She is a member of Xi'an Jiaotong University, Electronic Information and Engineering School, Founder of Xi'an Singularity One Information Technology (China) Co.,Ltd. Her currently research on the intersection of eye gaze tracking in human-computer interaction, psychology and human factors engineering, eye gaze tracking in the direction of image retrieval and image quality assessment, and research on eye-tracking software and hardware development and OEM products

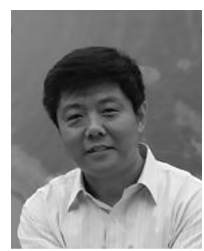

**Danmin Miao** is a Professor and Ph.D. supervisor with the Department of military medical psychology, Air Force Military Medical University, Chinese psychologist. Director of the National Psychological Testing Technology Center for conscription, director of the military mental health research center, director of the military Key Laboratory of medical psychology. He is currently engaged in military psychological selection and other fields.